\documentclass[11pt]{article}
\usepackage{coling2020}
\usepackage{times}
\usepackage{url}
\usepackage{latexsym}
\usepackage{multirow}
\usepackage{graphicx}
\usepackage{CJKutf8}
\usepackage{tabularx}

\usepackage{makecell}
\usepackage[utf8]{inputenc}
\usepackage[T1]{fontenc}

\newcommand{\korean}[1]{\begin{CJK*}{UTF8}{mj}{#1}\end{CJK*}}

\usepackage{amssymb}

\usepackage{amsmath}
\usepackage{multirow}

\usepackage{tipa}

\usepackage{color, colortbl}

\usepackage[table]{xcolor}
\definecolor{Gray}{gray}{0.9}

\title{KoParadigm: A Korean Conjugation Paradigm Generator}

\author{Kyubyong Park \\
  Kakao Brain \\
  {kyubyong.park@kakaobrain.com} \\
  }

\date{}

\begin{document}
\maketitle

\begin{abstract}
Korean is a morphologically rich language. Korean verbs change their forms in a fickle manner depending on tense, mood, speech level, meaning, etc. Therefore, it is challenging to construct comprehensive conjugation paradigms of Korean verbs. In this paper we introduce a Korean (verb) conjugation paradigm generator, dubbed \emph{KoParadigm}. To the best of our knowledge, it is the first Korean conjugation module that covers all contemporary Korean verbs and endings. KoParadigm is not only linguistically well established, but also computationally simple and efficient. We share it via PyPi.

\end{abstract}
\section{Introduction}
\label{intro}

%
%

A (morphological) paradigm is a set of word forms having a common root, e.g., `boy, boys', or `pretty, prettier, prettiest', or `look, looks, looked, looking'.
It has been of intense interest to both linguists and NLP researchers for a long time. Not only have paradigms been applied to education industry \cite{stracke2006,lester2009the,beyer2013}, but they also have been used widely for various NLP tasks such as morphological reinflections \cite{cotterell2016sigmorphon,cotterell2017conll,cotterell2018conll,mccarthy-etal-2019-sigmorphon}, lemmatizer generation \cite{nicolai-kondrak-2016-leveraging,bergmanis-goldwater-2019-training}, grammatical error correction (GEC) \cite{flachs2019noisy}, to name a few.

Korean is a morphologically rich language. To put it another way, a Korean word can have a lot of forms. In particular, Korean verbs change their forms colorfully according to their tense, mood, speech level, meaning, etc. For that reason construction of complete verb paradigms in the Korean language is well known to be challenging.

There have been many efforts to address it. 
Several academic books \cite{song1988,nam2007,kim2009oegugin,park2011,park2015} are designed to display the paradigms of a select few verbs in the Korean language. Although they all serve the pedagogical purpose, they may be less desirable for people who want to see the paradigms of a certain verb not in the list.

In addition to the books, there are a few online based multilingual conjugation tools. Among them are Verbix\footnote{\url{https://www.verbix.com/}} and UniMorph\footnote{\url{https://unimorph.github.io/}} \cite{kirov2016very,kirov2018unimorph}. Verbix shows paradigm tables of verbs for more than 300 languages including Korean. Admittedly, it is a valuable educational resource, but it has several limitations. It is far from complete, and contains a few critical errors\footnote{For example, it does not distinguish between action verbs and descriptive verbs. This causes many incorrect word forms. See Section \ref{4.3} for more.}. 
We also regret that the source code link\footnote{\url{https://xp-dev.com/svn/KoreanConjugator/}} provided at the web page seems to be broken.
UniMorph is a collection of inflectional paradigms primarily based on Wiktionary across many languages. Unfortunately, Korean is not included yet.

Motivated by the fact that there are no comprehensive and linguistically well established Korean verb paradigms, we develop a Korean (verb) conjugator in Python, dubbed \emph{KoParadigm}. We share all the source code and the package via GitHub\footnote{\url{https://github.com/kyubyong/koparadigm}} and PyPi\footnote{\url{https://pypi.org/project/KoParadigm}} respectively. To the best of our knowledge, it is the first Korean conjugation tool that comprehensively covers all contemporary Korean verbs and endings.  

\section{Glossary}
We believe it would be a good idea to provide the definitions of a few linguistic terms before we proceed as they will appear several times throughout the paper. 

\begin{itemize}
\item \textbf{Ending} An ending is a bound morpheme attached to a verb. For example, the \emph{s} in `boy\underline{s}' is a noun ending denoting plurality, and the \emph{ed} in `I walk\underline{ed}' is a verb ending forming the past tense. 

\item \textbf{Inflection} Inflection refers to a process of word formation in which a word is modified to express grammatical meanings. In the examples above, one can say ``\emph{boy} is inflected by adding the ending \emph{s} and \emph{walk} is inflected by adding the ending \emph{ed}''.

\item \textbf{Conjugation} The inflection of verbs is called conjugation. In other words, conjugation is a process of verb formation where a verb stem, the part that does not change during inflection, combines with one ore more endings. 

\item \textbf{Paradigm} A paradigm is a set of word forms having a common root. Accordingly, a conjugation paradigm is a set of verb forms from the same abstract lexical unit, or lexeme. For instance, the conjugation paradigm of \textsc{go} is `go, goes, went, gone, going'. 

\item \textbf{Action Verb \& Descriptive Verb} Korean verbs are divided into two groups: action verbs and descriptive verbs. Action verbs involve some action or movement, while descriptive verbs engage in the description of things. From the semantic perspective, the latter is close to adjectives, but unlike English adjectives, descriptive verbs have many syntactic properties as verbs.

\item \textbf{Light Vowel \& Dark Vowel} There are two kinds of vowels in Korean vowels. Light vowels, which correspond to the phonemic values  \textipa{/a, o, 2/}, deliver some nuance of being small, light, bright, etc., while dark vowels convey the opposite values.

\end{itemize}

\section{\emph{Why} is it Hard to Generate Korean Conjugation Paradigms and \emph{How} do we Take On Such Challenges?}

We state firmly that it is challenging to make a conjugation paradigm generator for Korean verbs especially if one seeks perfection and completeness. Why is it so?

Firstly, the conditions in which verbs and endings conjoin are complicated. A certain type of endings combine with certain types of verbs exclusively. This will be further discussed in Section \ref{4.2}.
Secondly, there are many conjugation patterns Korean verbs follow, and there are also many exceptions that refuse to do so. Some of them are easily identifiable, but some others are not.
Lastly, it is never easy to design a conjugation algorithm that is simple enough for humans to understand and at the same time efficient enough for machines to process.
Letters in Hangul, the Korean writing system, are classified into vowels and consonants, like English alphabets. However, Hangul is different from English alphabets in that in Hangul a vowel and one or more consonants gather together to form a syllable, which is an independent character. Because the conjugation process occurs at the letter level rather than the syllable level, we need to think how effectively we apply such Hangul characteristic to the conjugation algorithm.

We take on each challenge like the following.
First, we carefully examine the conditions where verbs and endings combine using an authoritative Korean dictionary and classify them systematically by binary features. 
Second, we classify every contemporary Korean verb by its conjugation pattern based on the dictionary and other reference books \cite{song1988,nam2007,kim2009oegugin,park2011,park2015}. 
Finally, we design a conjugation template in which a set of combination rules between verb classes and ending classes are arranged in a tabular format. From the associated rules in the template, the paradigms of a verb are generated dynamically.  
We find that this approach is advantageous for the overall process of our project as we are to write, modify, and test the rules frequently throughout the project period.

\section{Method}

In this section we explain how we develop our verb conjugator. We first draw up the lists of Korean endings and verbs (Section \ref{4.1}). Next, we classify the endings (Section \ref{4.2}) and the verbs (Section \ref{4.3}).
Finally, we construct a conjugation template, a set of rules where verbs and endings join to form word forms (Section \ref{4.4}).

    \subsection{Lists of Endings and Verbs}
\label{4.1}
We collect the lists of Korean endings and verbs from \emph{Standard Korean Language Dictionary}\footnote{\url{https://stdict.korean.go.kr}}. It has been considered the \emph{de facto} standard as the name directly claims since it was published in 1999. On March 11, 2019, it was officially announced\footnote{\url{https://stdict.korean.go.kr/notice/noticeList.do}} that the license of the dictionary changed to `CC BY-SA 2.0 KR', so it became possible for the public to freely download the digital contents.

We discard any entry if it is deprecated or used regionally because we do not have enough linguistic knowledge of such items. 
The resulting data consists of 608 endings and 73,759 verbs.

    \subsection{Ending Classification}
\label{4.2}
Subsequently, we classify the collected Korean endings into 24 categories by the 14 binary distinctive features (\textbf{A}-\textbf{N}) in Table \ref{tab:ending_classes}. 
We emphasize that it is important to work on ending classification prior to verb classification. 
When a verb and an ending join, the former seems to 
\emph{choose} the latter on first thought because the ending is attached to its preceding verb. However, indeed the opposite is true; endings are picky to choose their partner verbs. For example, the dictionary describes that the ending \korean{느냐} nunya `interrogative' can be combined with action verbs as well as \korean{있}, \korean{없}, and \korean{계시}. On the contrary, there is no information about which kinds of endings those verbs can combine with in the dictionary. Accordingly, we first identify the conditions where endings combine with verbs, and then reuse them to classify verbs.

%

\begin{table}
\renewcommand{\arraystretch}{1.0}
\rowcolors{1}{}{Gray}
\begin{tabular}{c|p{0.1cm}p{0.1cm}p{0.1cm}p{0.1cm}p{0.1cm}p{0.1cm}p{0.1cm}p{0.1cm}p{0.1cm}p{0.1cm}p{0.1cm}p{0.1cm}p{0.1cm}c|c|l}

\Xhline{1.0pt}

Ending  & \multicolumn{14}{c|}{Features} &\# End-& \multirow{2}{*}{Examples} \\
\cline{2-15}
Class & A & B & C & D & E & F & G & H & I & J & K & L & M & N &  ings  &   
\\

\Xhline{1.0pt}

\rowcolors{1}{}{Gray}
1 & \checkmark & \checkmark & \checkmark & \checkmark & \checkmark & \checkmark & \checkmark & \checkmark & \checkmark & \checkmark & \checkmark & \checkmark & \checkmark &  & 77 &  \korean{거나} kena `or'                     
\\
2 & \checkmark & \checkmark & \checkmark & \checkmark & \checkmark &  & \checkmark &  & \checkmark & \checkmark & \checkmark &  & \checkmark &  & 86 &  \korean{니까} nikka `thus'   
\\
3 & \checkmark & \checkmark & \checkmark & \checkmark &  & \checkmark & \checkmark & \checkmark &  &  & \checkmark & \checkmark & \checkmark & \checkmark & 11 &  \korean{어야} eya `if'  
\\
4 & \checkmark & \checkmark &  & \checkmark & \checkmark & \checkmark & \checkmark & \checkmark & \checkmark & \checkmark & \checkmark & \checkmark & \checkmark &  & 6 &  \korean{고는} konun `if'                    
\\
5 & \checkmark & \checkmark &  & \checkmark & \checkmark &  & \checkmark &  & \checkmark & \checkmark & \checkmark &  & \checkmark &  & 24 &  \korean{리만치} limanchi `as ... as'        
\\
6 & \checkmark & \checkmark &  &  & \checkmark &  &  &  &  & \checkmark &  & \checkmark & \checkmark & \checkmark & 12 &  \korean{아야} aya `if'                   
\\
7 & \checkmark & \checkmark &  &  &  & \checkmark &  & \checkmark &  &  &  & \checkmark &  & \checkmark & 95 &  \korean{으나} una `but'
\\
8 & \checkmark & \checkmark &  &  &  & \checkmark &  & \checkmark &  &  &  & \checkmark &  &  & 9 &  \korean{습니다} supnita `present'
\\
9 & \checkmark &  &  &  & \checkmark & \checkmark & \checkmark &  & \checkmark & \checkmark & \checkmark & \checkmark & \checkmark &  & 36 &  \korean{건대} kentay `according to'
\\
10 & \checkmark &  &  &  & \checkmark &  & \checkmark &  & \checkmark & \checkmark & \checkmark &  & \checkmark &  & 50 &  \korean{십시오} `imperative'  
\\
11 & \checkmark &  &  &  & \checkmark &  &  &  & \checkmark & \checkmark &  & \checkmark & \checkmark & \checkmark & 2 &  \korean{아다} ata `then'    
\\
12 & \checkmark &  &  &  &  & \checkmark & \checkmark & \checkmark &  & \checkmark & \checkmark & \checkmark & \checkmark &  & 25 &  \korean{느냐} nunya `interrogative'         
\\
13 & \checkmark &  &  &  &  & \checkmark &  &  &  &  &  & \checkmark &  & \checkmark & 42 &  \korean{으라고} ulako `indirect speech'  
\\
14 & \checkmark &  &  &  &  & \checkmark &  &  &  &  &  & \checkmark &  &  & 22 &  \korean{는다} nunta `present'    
\\
15 & \checkmark &  &  &  &  &  &  &  &  &  & \checkmark & \checkmark & \checkmark & \checkmark & 2 &  \korean{어다} eta `then'    
\\
16 &  & \checkmark & \checkmark & \checkmark &  & \checkmark & \checkmark & \checkmark &  & \checkmark & \checkmark & \checkmark & \checkmark &  & 3 &  \korean{구나} kwuna `exclamatory' 
\\
17 &  & \checkmark & \checkmark & \checkmark &  & \checkmark & \checkmark &  &  & \checkmark & \checkmark &  &  &  & 2 &  \korean{뇨} nyo `interrogative'             
\\
18 &  & \checkmark & \checkmark & \checkmark &  &  & \checkmark &  &  & \checkmark & \checkmark &  & \checkmark &  & 10 &  \korean{ㄴ가} ka `interrogative'            
\\
19 &  & \checkmark &  & \checkmark &  & \checkmark & \checkmark & \checkmark &  & \checkmark & \checkmark & \checkmark & \checkmark &  & 22 &  \korean{다더라} tatela `reportedly'           
\\
20 &  & \checkmark &  &  &  &  &  &  &  &  &  & \checkmark &  & \checkmark & 14 &  \korean{으냐} unya `interrogative'           
\\
21 &  &  & \checkmark & \checkmark &  &  &  &  &  & \checkmark & \checkmark &  &  & \checkmark & 1 &  \korean{에요} eyyo `declarative'             
\\
22 &  &  & \checkmark & \checkmark &  &  &  &  &  & \checkmark & \checkmark &  &  &  & 44 &  \korean{로구려} lokwulye `exclamatory'     
\\
23 &  &  &  &  & \checkmark &  &  &  &  & \checkmark & \checkmark &  &  &  & 2 &  \korean{너라} nela `imperative'          
\\
24 &  &  &  &  &  &  &  &  & \checkmark & \checkmark & \checkmark &  &  & \checkmark & 11 &  \korean{여야} yeya `if'   
\\

\Xhline{1.0pt}
\end{tabular}

\caption{Ending classes. \textbf{A}: FollowedByActionVerb, \textbf{B}: FollowedByDescriptiveVerb, \textbf{C}: FollowedBy\korean{이}, \textbf{D}: FollowedBy\korean{아니}, \textbf{E}: FollowedBy\korean{오}, \textbf{F}: FollowedBy\korean{있}, \textbf{G}: FollowedBy\korean{계시}, \textbf{H}: FollowedBy\korean{없}, \textbf{I}: FollowedBy\korean{하}, \textbf{J}: FollowedBySyllableWithLightVowel, \textbf{K}: FollowedBySyllableWithDarkVowel, \textbf{L}: FollowedByAnyConsonantBut\korean{ㄹ}, \textbf{M}: FollowedBy\korean{ㄹ}, \textbf{N}: StartsWithVowel.} 
\label{tab:ending_classes}
\end{table}

\renewcommand{\labelenumi}{\Alph{enumi}.}
\begin{enumerate}

    \item[\textbf{A-B.}] \textbf{FollowedBy\{ActionVerb,DescriptiveVerb\}} Many endings discriminate between action verbs and descriptive verbs. In other words, many endings can combine with either action verbs \emph{or} descriptive verbs, but not both. See Classes 9-20.

     \item[\textbf{C-I.}] \textbf{FollowedBy\{\korean{이,아니,오,있,계시,없,하}\}} There is no rule without an exception. And the Korean language is no exception, of course. Korean verbs \korean{이, 아니, 오, 있, 계시, 없}, and \korean{하} deviate from regular conjugation patterns, thus need to be treated as special. Some endings are sensitive to them.

      \item[\textbf{J-K.}] \textbf{FollowedBySyllableWith\{LightVowel,DarkVowel\}} Vowel harmony is present in the Korean language. In Korean, vowels \korean{ㅏ, ㅗ, ㅑ, ㅛ, ㅘ, ㅚ} and \korean{ㅐ} are classified as light vowels, and the rest dark vowels. The vowels in each group tend to cluster together. Therefore, most endings that begin with a light vowel prefer verbs that end in a syllable with a light vowel than a dark vowel. Compare Classes 3 and 6.

     \item[\textbf{L-M.}] \textbf{FollowedBy\{AnyConsonantBut\korean{ㄹ},\korean{ㄹ}\}} Some endings are attached to the verbs ending in \korean{ㄹ}, but not to the verbs ending in one of any other consonants, and some other endings do the other way round.
     Compare Classes 5 and 7.

     \item[\textbf{N.}] \textbf{StartsWithVowel} When an ending starting with a vowel is immediately followed by a vowel, different types of vowel contraction can occur.

\end{enumerate}

\begin{table}
\renewcommand{\arraystretch}{0.9}
\rowcolors{1}{}{Gray}
\begin{tabular}{c|cccccccccccc|c|l}

\Xhline{1.0pt}

Verb & \multicolumn{12}{c|}{Features} & \multirow{2}{*}{\# Verbs} & \multirow{2}{*}{Examples} 
\\
\cline{2-13}
                                  
Class& A &B&C&D&E&F&G&H&I&J&K&L& &                                        \\
\Xhline{1.0pt}

1&\checkmark&\checkmark&\checkmark&\checkmark&&&&&&&&&13& \korean{있} iss `exist'         \\
2&\checkmark&\checkmark&\checkmark&&&&&&&&&&1& \korean{계시} kyeysi `stay'      \\
3&\checkmark&\checkmark&&&&\checkmark&&&&&&\checkmark&37216& \korean{원하} wenha `want'       \\
4&\checkmark&\checkmark&&&&&&&&&&\checkmark&73& \korean{오} o `come'            \\
5&\checkmark&\checkmark&&&&&&&&&&&15& \korean{그러} kule `do so'       \\
6&\checkmark&\checkmark&&&&&&&&\checkmark&&&1& \korean{이르} ilu `reach'        \\
7&\checkmark&&\checkmark&\checkmark&&&&&&&&&151& \korean{없} eps `not exist'     \\
8&\checkmark&&\checkmark&\checkmark&&&&&&&\checkmark&&1& \korean{그렇} kuleh `yes'        \\
9&\checkmark&&\checkmark&&&\checkmark&&&&&&\checkmark&14615& \korean{강하} kangha `strong'    \\
10&\checkmark&&\checkmark&&&&&&&&&&1& \korean{아니} ani `no'           \\
11&\checkmark&&\checkmark&&&&&&&&&&13& \korean{아프} aphu `sick'        \\
12&\checkmark&&\checkmark&&&&&&&&&&29& \korean{기쁘} kippu `happy'      \\
13&\checkmark&&\checkmark&&&&&&&\checkmark&&&17& \korean{푸르} phwulu `azure'     \\
14&\checkmark&&&&&&&&&&&&1& \korean{이} i `be'  \\
15&&\checkmark&&\checkmark&\checkmark&&&&&&&\checkmark&132& \korean{알} al `know'           \\
16&&\checkmark&&\checkmark&\checkmark&&&&&&&&365& \korean{열} yel `open'          \\
17&&\checkmark&&\checkmark&&&&&&&&\checkmark&496& \korean{막} mak `defend'         \\
18&&\checkmark&&\checkmark&&&&&&&&&452& \korean{먹} mek `eat'         \\
19&&\checkmark&&\checkmark&&&\checkmark&&&&&\checkmark&2& \korean{깨닫} kkaytat `realize'  \\
20&&\checkmark&&\checkmark&&&\checkmark&&&&&&38& \korean{듣} tut `hear'          \\
21&&\checkmark&&\checkmark&&&&\checkmark&&&&\checkmark&1& \korean{돕} top `help'          \\
22&&\checkmark&&\checkmark&&&&\checkmark&&&&&20& \korean{눕} nwup `lie'          \\
23&&\checkmark&&\checkmark&&&&&\checkmark&&&\checkmark&2& \korean{낫} nas `recover'       \\
24&&\checkmark&&\checkmark&&&&&\checkmark&&&&55& \korean{긋} kus `draw'          \\
25&&\checkmark&&\checkmark&&&&&&\checkmark&&&58& \korean{고르} kolu `choose'      \\
26&&\checkmark&&\checkmark&&&&&&\checkmark&&&141& \korean{부르} pwulu `call'       \\
27&&\checkmark&&&&&&&&&&\checkmark&525& \korean{가} ka `go'             \\
28&&\checkmark&&&&&&&&&&&16835& \korean{주} cwu `give'          \\
29&&\checkmark&&&&&&&&&&&15& \korean{담그} tamku `soak'       \\
30&&\checkmark&&&&&&&&&&&96& \korean{끄} kku `put out'       \\
31&&&\checkmark&\checkmark&\checkmark&&&&&&&\checkmark&12& \korean{달} tal `sweet'         \\
32&&&\checkmark&\checkmark&&&&&&&&\checkmark&195& \korean{작} cak `small'         \\
33&&&\checkmark&\checkmark&&&&&&&&&27& \korean{희} huy `white'         \\
34&&&\checkmark&\checkmark&&&&\checkmark&&&&\checkmark&1& \korean{곱} kop `pretty'        \\
35&&&\checkmark&\checkmark&&&&\checkmark&&&&&1376& \korean{덥} tep `hot'           \\
36&&&\checkmark&\checkmark&&&&&\checkmark&&&\checkmark&1& \korean{낫} nas `better'        \\
37&&&\checkmark&\checkmark&&&&&&&\checkmark&\checkmark&64& \korean{빨갛} ppalkah `red'      \\
38&&&\checkmark&\checkmark&&&&&&&\checkmark&\checkmark&5& \korean{하얗} hayah `white'      \\
39&&&\checkmark&\checkmark&&&&&&&\checkmark&&39& \korean{뻘겋} ppelkeh `red'      \\
40&&&\checkmark&\checkmark&&&&&&&\checkmark&&9& \korean{허옇} heyeh `white'      \\
41&&&\checkmark&&\checkmark&&&&&&&&56& \korean{서툴} sethwul `poor'     \\
42&&&\checkmark&&&&&&&&&\checkmark&86& \korean{힘차} himcha `energetic' 
\\
43&&&\checkmark&&&&&&&&&&305& \korean{멋지} mesci `gorgeous'   
\\
44&&&\checkmark&&&&&&&&&&137& \korean{젊} celm `young'        \\
45&&&\checkmark&&&&&&&\checkmark&&&49& \korean{다르} talu `different'   \\
46&&&\checkmark&&&&&&&\checkmark&&&19& \korean{부르} pwulu `full'       \\

\Xhline{1.0pt}
\end{tabular}
\caption{Verb classes. \textbf{A}: IsSpecial, \textbf{B}: IsActionVerb, \textbf{C}: IsDescriptiveVerb, \textbf{D}: EndsWithConsonant, \\ \textbf{E}: EndsWith\korean{ㄹ}, \textbf{F}: EndsWith\korean{하}, \textbf{G}: EndsWith\korean{ㄷ}, \textbf{H}: EndsWith\korean{ㅂ}, \textbf{I}: EndsWith\korean{ㅅ}, \textbf{J}: EndsWith\korean{르}, \textbf{K}: EndsWith\korean{ㅎ}, \textbf{L}: EndsInSyllableWithLightVowel
} 

\label{tab:verb_classes}

\end{table}


    \subsection{Verb Classification}
\label{4.3}

Next, we classify the collected Korean verbs into 46 categories based the 12 binary features (\textbf{A}-\textbf{L}), as shown in Table \ref{tab:verb_classes}. 

\renewcommand{\labelenumi}{\Alph{enumi}.}

\begin{enumerate}

\item[\textbf{A.}] \textbf{IsSpecial} Classes 1-14 are special groups as they do not follow the regular patterns.

\item[\textbf{B--C}.]  \textbf{Is\{ActionVerb,DescriptiveVerb\}} Action verbs and descriptive verbs in the Korean language behave differently in terms of conjugation. Therefore, we check if a verb is an action verb and/or a descriptive verb.
The reason we add both features is that two verbs, \korean{있} iss `exist` and \korean{계시} kyeysi `stay', can be classified into both. See Classes 1 and 2.

\item[\textbf{D}.] \textbf{EndsWithConsonant} We check if a verb ends with a consonant or a vowel. As mentioned earlier, if a verb that ends with a vowel is combined with a vowel, those vowels can be contracted. For example, when the verb \korean{주} cwu `give' combines with \korean{어라} ela `imperative' to form \korean{주어라}, the final \korean{ㅜ} in the verb and the vowel \korean{ㅓ} in the ending can be shortened to \korean{ㅝ}, thereby yielding \korean{줘라} as an alternative form.

\item[\textbf{E--K}.]  \textbf{EndsWith\{\korean{ㄹ,하,ㄷ,ㅂ,ㅅ,르,ㅎ}\}} The conjugation patterns of Korean verbs differ according to their final consonants or syllables. We check if a verb ends with \korean{ㄹ, 하, ㄷ, ㅂ, ㅅ, 르}, and \korean{ㅎ}.

\item[\textbf{L}.] \textbf{EndsInSyllableWithLightVowel} This feature is related to the vowel harmony we briefly explained in Section \ref{4.2} \textbf{J-K}.

\end{enumerate}

    \subsection{Conjugation Template}
\label{4.4}

Finally, with the 24 ending classes and 46 verb classes we create a conjugation template designed to show the combination rules between each verb class (row) and ending class (column) in a tabular form.
We provide its first 10 verb classes $\times$ 6 ending classes in Table \ref{tab:template} for illustration. (The complete template is available in Appendix \ref{appendix:1}.)
In Table \ref{tab:template}, some cells (e.g., Verb Class 1 $\times$ Ending Class 2) are blank, which means that the associated verb classes and ending classes do not conjoin. Otherwise, each rule has its own rule---three elements separated by commas. For example, the rule of Verb Class 1 $\times$ Ending Class 1 is ``None,,None'', and that of Verb Class 8 $\times$ Ending Class 3 is ``-2,\korean{ㅐ},2''. 
\begin{table}[ht]
\centering
\renewcommand{\arraystretch}{1.1}
\begin{tabular}{c|cccccc}
\Xhline{1.0pt}

\multirow{2}{*}{Verb Class} & \multicolumn{6}{c}{Ending Class} \\
 \cline{2-7}
 & 1&2&3&4&5&6\\

\Xhline{1.0pt}

1&None,,None&&None,,None&None,,None&& \\
2&None,,None&None,,None&None,,None&None,,None&None,,None& \\
3&None,,None&None,,None&&None,,None&None,,None& \\
4&None,,None&None,,None&&None,,None&None,,None&None,,1 \\
5&None,,None&None,,None&-1,\korean{ㅐ},2&None,,None&None,,None& \\
6&None,,None&None,,None&None,\korean{ㄹ},1&None,,None&None,,None& \\
7&None,,None&&None,,None&None,,None&& \\
8&None,,None&&-2,\korean{ㅐ},2&None,,None&& \\
9&None,,None&None,,None&&None,,None&None,,None& \\
10&None,,None&None,,None&None,,None&None,,None&None,,None& \\
\Xhline{1.0pt}
\end{tabular}
\caption{Conjugation template. Note that only some rows (verb classes) and columns (ending classes) are shown due to space constraints. The complete table is available in Appendix 
.}
\label{tab:template}
\end{table}
\begin{figure}[ht]
    \centering
    \includegraphics[width=15.5cm,clip=false]{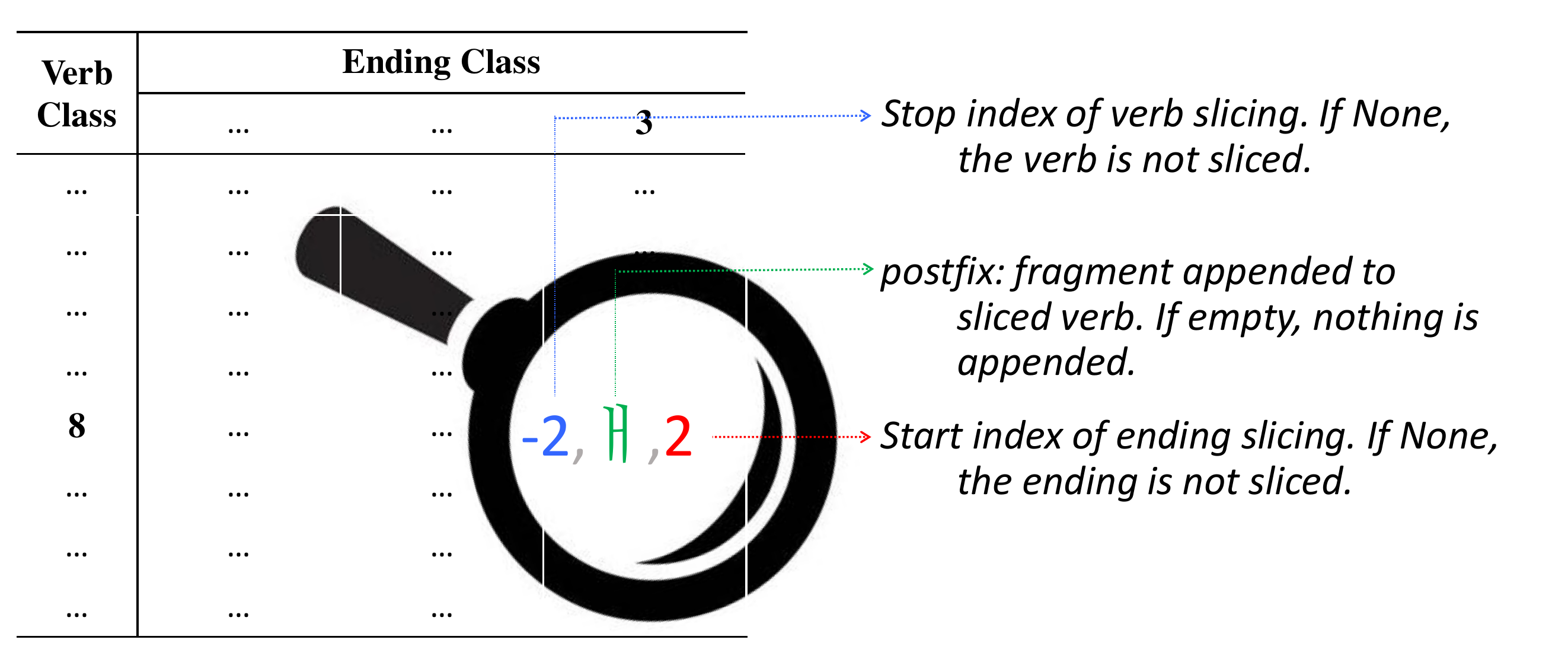}
    \caption{Rule notation.}
    \label{fig:rule}
\end{figure}

 
The meanings of the three elements are illustrated by the example of Verb Class 8 $\times$ Ending Class 3 in Figure \ref{fig:rule}. The first one is the stop index of the verb slicing. It is either a negative integer, which means ``Drop the \textbf{last} \emph{n} letters of the verb'', or None, which means ``No slicing''. The second element is appended to the sliced verb. Of course, nothing is appended if it is empty. And, the last element is the start index of the ending slicing.  It is either a positive integer, which means ``Drop the \textbf{first} \emph{n} letters of the ending'', or None, which means ``No slicing''.

Now that we have the verb (classes), ending (classes), and their combination rule, we are ready to combine them to generate a new word form. 
We illustrate the process step by step. We assume that we are given the verb \korean{그렇} in Verb Class 8, the ending \korean{어야} in Ending Class 3, and their combination rule ``-2,\korean{ㅐ},2''.

\renewcommand{\labelenumi}{\arabic{enumi}.}
\begin{itemize}

    \item \textbf{STEP 1.} Decompose the verb \korean{그렇} into a sequence of letters, \korean{ㄱㅡㄹㅓㅎ}. 
    
    \item \textbf{STEP 2.} Slice the letters with the stop index -2. As a result of \korean{ㄱㅡㄹㅓㅎ}[:-2], \korean{ㄱㅡㄹ} remains. 
    
     \item \textbf{STEP 3.} Merge the sliced verb letters and the postfix, the second element of the rule. We have \korean{ㄱㅡㄹㅐ} after \korean{ㄱㅡㄹ}+\korean{ㅐ}.
     
     \item \textbf{STEP 4.} Decompose the ending \korean{어야} into a sequence of letters, \korean{ㅇㅓㅇㅑ}. 

     \item \textbf{STEP 5.} Slice the letters with the start index 2. As a result of \korean{ㅇㅓㅇㅑ}[2:], \korean{ㅇㅑ} remains. 
     
      \item \textbf{STEP 6.} Merge the outputs of \textbf{STEP 3} and \textbf{STEP 5}. We have \korean{ㄱㅡㄹㅐㅇㅑ} after \korean{ㄱㅡㄹㅐ}+\korean{ㅇㅑ}.

      \item \textbf{STEP 7.} Compose the result into Hangul syllables. Finally, \korean{그래야} is returned.
      
\end{itemize}

\section{KoParadigm: A Python Package of Korean Conjugation Paradigm Generation}

In order to make our work easily accessible, we package it under the name of \emph{KoParadigm}. 
It is available on PyPi as \url{https://pypi.org/project/KoParadigm}.

\subsection{Implementation}

The package is extremely simple; it consists of a resource file---\texttt{koparadigm.xlsx}---and a Python code file---\texttt{koparadigm.py}.
The \texttt{koparadigm.xlsx} file contains three sheets: \texttt{Endings}, \texttt{Verbs}, \texttt{Template}. 
In the first two sheets all endings and verbs along with their class ids are written, and the last one has the conjugation template. 
On the other hand, \texttt{koparadigm.py} has the main Python class \texttt{Paradigm} and a utility function \texttt{prettify}, which helps pretty-print the paradigms.

\renewcommand{\labelitemii}{$\bullet$}

\subsection{Usage}

\begin{figure}[ht]
    \centering
    \includegraphics[width=15.5cm,clip=false]{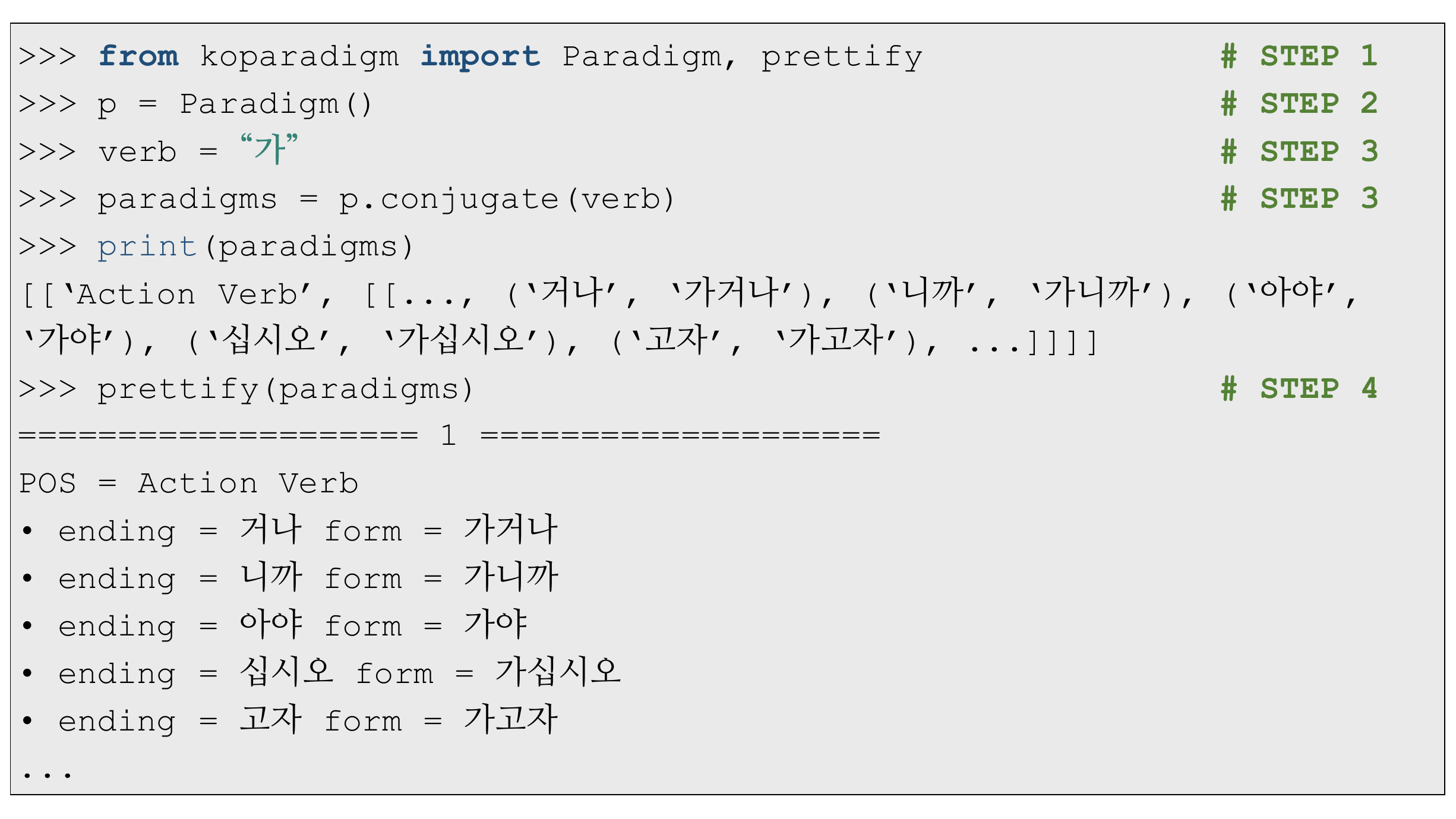}
    \caption{KoParadigm usage example.}
    \label{fig:code}
\end{figure}

KoParadigm provides an easy-to-use Python interface. We explain how to use it step by step with an authentic example in Figure \ref{fig:code}.

\renewcommand{\labelenumi}{\arabic{enumi}.}
\begin{itemize}

    \item \textbf{STEP 1.} Import \texttt{Paradigm} and a custom utility function \texttt{prettify}. 
    
    \item \textbf{STEP 2.} Instantiate the Python class \texttt{Paradigm}. It makes three Python dictionaries: \texttt{verb2verb\_classes}, \texttt{ending\_class2endings}, \texttt{verb\_class2rules}. Note that \texttt{rules} is a list of tuples consisting of an ending and a combination rule.

     \item \textbf{STEP 3.} Get the paradigms of the query verb with the method \texttt{conjugate}. Under the hood, the following happen.
     
     \begin{itemize}
      \item \textbf{STEP 3-1.} Look up the query in \texttt{verb2verb\_classes}.
      
      \item \textbf{STEP 3-2.} If found, retrieve the class ids of the verb. Otherwise, return ``Not Found".
      
      \item \textbf{STEP 3-3.} Search the \texttt{verb\_class2rules} for each verb class id and retrieve the associated ending classes and rules.
      
      \item \textbf{STEP 3-4.} Fetch the endings that belong to each of the ending class ids from the \texttt{ending\_class2ending}.
      
      \item \textbf{STEP 3-5.} Generate the word form one by one via the combination process explained in Section \ref{4.4}.

     \end{itemize}
     
     \item \textbf{STEP 4.} Print the paradigms in the neat format using \texttt{prettify}. 
                                                        
\end{itemize}

\section{Conclusion and Future Work}

In this paper, we introduced why and how we developed KoParadigm, a new Korean conjugation paradigm generator.
Not only is it linguistically well established, but also it provides comprehensive conjugation paradigms of Korean verbs in a computationally simple manner.

We would like to improve it in the future. 
First of all, we will make efforts to find and fix any errors. Although we did our best, we humbly admit that there may be some mistakes we are unaware of.
And, we plan to add useful glosses such as the meanings of verbs and endings.
Lastly, we believe it would help second language learners if we could show Romanization next to Hangul.

\section*{Acknowledgements}
We thank Jiyeon Ham and Sungjoon Park for their helpful discussions.


\bibliographystyle{acl}
\bibliography{coling2020}

\section{Appendix}
\appendix

    \section{Complete Conjugation Template}
\label{appendix:1}
\clearpage
\begin{table}[ht]
\renewcommand{\arraystretch}{1.2}
\small
\rowcolors{1}{}{Gray}
\begin{tabular}{c|ccccccc}
\Xhline{1.0pt} 
\rowcolors{1}{}{Gray}
Ending Class  $\triangleright$  &      \multirow{2}{*}{1}         & \multirow{2}{*}{2}    & \multirow{2}{*}{3}         & \multirow{2}{*}{4}         & \multirow{2}{*}{5}         & \multirow{2}{*}{6}    &
\multirow{2}{*}{7}
 \\

Verb Class $\triangledown$&&&&&&& \\
\Xhline{1.0pt}

1&None,,None&&None,,None&None,,None&&&None,,None \\
2&None,,None&None,,None&None,,None&None,,None&None,,None&& \\
3&None,,None&None,,None&&None,,None&None,,None&& \\
4&None,,None&None,,None&&None,,None&None,,None&None,,1& \\
5&None,,None&None,,None&-1,\korean{ㅐ},2&None,,None&None,,None&& \\
6&None,,None&None,,None&None,\korean{ㄹ},1&None,,None&None,,None&& \\
7&None,,None&&None,,None&None,,None&&&None,,None \\
8&None,,None&&-2,\korean{ㅐ},2&None,,None&&&-1,,2 \\
9&None,,None&None,,None&&None,,None&None,,None&& \\
10&None,,None&None,,None&None,,None&None,,None&None,,None&& \\
11&None,,None&None,,None&-1,\korean{ㅏ},2&None,,None&None,,None&& \\
12&None,,None&None,,None&-1,,1&None,,None&None,,None&& \\
13&None,,None&None,,None&None,\korean{ㄹ},1&None,,None&None,,None&& \\
14&None,,None&None,,None&None,,None&&&& \\
15&None,,None&-1,,None&&None,,None&-1,,None&None,,None& \\
16&None,,None&-1,,None&None,,None&None,,None&-1,,None&& \\
17&None,,None&&&None,,None&&None,,None&None,,None \\
18&None,,None&&None,,None&None,,None&&&None,,None \\
19&None,,None&&&None,,None&&-1,\korean{ㄹ},None&-1,\korean{ㄹ},None \\
20&None,,None&&-1,\korean{ㄹ},None&None,,None&&&-1,\korean{ㄹ},None \\
21&None,,None&&&None,,None&&-1,\korean{ㅇㅘ},2&-1,\korean{ㅇㅜ},2 \\
22&None,,None&&-1,\korean{ㅇㅝ},2&None,,None&&&-1,\korean{ㅇㅜ},2 \\
23&None,,None&&&None,,None&&-1,,None&-1,,None \\
24&None,,None&&-1,,None&None,,None&&&-1,,None \\
25&None,,None&None,,None&&None,,None&None,,None&& \\
26&None,,None&None,,None&-2,\korean{ㄹㄹ},1&None,,None&None,,None&& \\
27&None,,None&None,,None&&None,,None&None,,None&None,,1& \\
28&None,,None&None,,None&None,,1&None,,None&None,,None&& \\
29&None,,None&None,,None&-1,\korean{ㅏ},2&None,,None&None,,None&& \\
30&None,,None&None,,None&-1,,1&None,,None&None,,None&& \\
31&None,,None&-1,,None&&None,,None&-1,,None&None,,None& \\
32&None,,None&&&None,,None&&None,,None&None,,None \\
33&None,,None&None,,None&None,,None&None,,None&None,,None&& \\
34&None,,None&&&None,,None&&-1,\korean{ㅇㅘ},2&-1,\korean{ㅇㅜ},2 \\
35&None,,None&&-1,\korean{ㅇㅝ},2&None,,None&&&-1,\korean{ㅇㅜ},2 \\
36&None,,None&&&None,,None&&-1,,None&-1,,None \\
37&None,,None&&&None,,None&&-2,\korean{ㅐ},2&-1,,2 \\
38&None,,None&&&None,,None&&-2,\korean{ㅒ},2&-1,,2 \\
39&None,,None&&-2,\korean{ㅔ},2&None,,None&&&-1,,2 \\
40&None,,None&&-2,\korean{ㅖ},2&None,,None&&&-1,,2 \\
41&None,,None&-1,,None&None,,None&None,,None&-1,,None&& \\
42&None,,None&None,,None&&None,,None&None,,None&None,,None& \\
43&None,,None&None,,None&None,,None&None,,None&None,,None&& \\
44&None,,None&&None,,None&None,,None&&&None,,None \\
45&None,,None&None,,None&-2,\korean{ㄹㄹㅏ},2&None,,None&None,,None&& \\
46&None,,None&None,,None&-2,\korean{ㄹㄹ},1&None,,None&None,,None&& \\

\Xhline{1.0pt}
\end{tabular}
\end{table}

\clearpage

\begin{table}[ht]
\renewcommand{\arraystretch}{1.2}
\small

\rowcolors{1}{}{Gray}
\begin{tabular}{c|ccccccc}
\Xhline{1.0pt} 
\rowcolors{1}{}{Gray}
Ending Class  $\triangleright$         & \multirow{2}{*}{8}         & \multirow{2}{*}{9}         & \multirow{2}{*}{10}        & \multirow{2}{*}{11}   & \multirow{2}{*}{12} & \multirow{2}{*}{13} &\multirow{2}{*}{14}

 \\
Verb Class $\triangledown$&&&&&&&
\\
\Xhline{1.0pt}

1&None,,None&None,,None&&&None,,None&None,,None&None,,None \\
2&&None,,None&None,,None&&None,,None&& \\
3&&None,,None&None,,None&&None,,None&& \\
4&&None,,None&None,,None&None,,1&None,,None&& \\
5&&None,,None&None,,None&&None,,None&& \\
6&&None,,None&None,,None&&None,,None&& \\
7&None,,None&&&&None,,None&& \\
8&None,,None&&&&&& \\
9&&&&&&& \\
10&&&&&&& \\
11&&&&&&& \\
12&&&&&&& \\
13&&&&&&& \\
14&&&&&&& \\
15&&None,,None&-1,,None&None,,None&None,,None&& \\
16&&None,,None&-1,,None&&None,,None&& \\
17&None,,None&None,,None&&None,,None&None,,None&None,,None&None,,None \\
18&None,,None&None,,None&&&None,,None&None,,None&None,,None \\
19&None,,None&None,,None&&-1,\korean{ㄹ},None&None,,None&-1,\korean{ㄹ},None&None,,None \\
20&None,,None&None,,None&&&None,,None&-1,\korean{ㄹ},None&None,,None \\
21&None,,None&None,,None&&-1,\korean{ㅇㅘ},2&None,,None&-1,\korean{ㅇㅜ},2&None,,None \\
22&None,,None&None,,None&&&None,,None&-1,\korean{ㅇㅜ},2&None,,None \\
23&None,,None&None,,None&&-1,,None&None,,None&-1,,None&None,,None \\
24&None,,None&None,,None&&&None,,None&-1,,None&None,,None \\
25&&None,,None&None,,None&&None,,None&& \\
26&&None,,None&None,,None&&None,,None&& \\
27&&None,,None&None,,None&None,,1&None,,None&& \\
28&&None,,None&None,,None&&None,,None&& \\
29&&None,,None&None,,None&&None,,None&& \\
30&&None,,None&None,,None&&None,,None&& \\
31&&&&&&& \\
32&None,,None&&&&&& \\
33&&&&&&& \\
34&None,,None&&&&&& \\
35&None,,None&&&&&& \\
36&None,,None&&&&&& \\
37&None,,None&&&&&& \\
38&None,,None&&&&&& \\
39&None,,None&&&&&& \\
40&None,,None&&&&&& \\
41&&&&&&& \\
42&&&&&&& \\
43&&&&&&& \\
44&None,,None&&&&&& \\
45&&&&&&& \\
46&&&&&&& \\
\Xhline{1.0pt}
\end{tabular}
\end{table}

\begin{table}[ht]
\renewcommand{\arraystretch}{1.2}
\small
\rowcolors{1}{}{Gray}
\begin{tabular}{c|ccccccc}
\Xhline{1.0pt} 
\rowcolors{1}{}{Gray}
Ending Class  $\triangleright$  & \multirow{2}{*}{15}     & \multirow{2}{*}{16}   & \multirow{2}{*}{17}     & \multirow{2}{*}{18}    & \multirow{2}{*}{19}     & \multirow{2}{*}{20}  &
 \multirow{2}{*}{21}
 \\
Verb Class $\triangledown$&&&&&&&
\\
\Xhline{1.0pt}

1&&None,,None&&&&& \\
2&&None,,None&None,,None&None,,None&&& \\
3&&&&&&& \\
4&&&&&&& \\
5&None,,2&&&None,,None&&& \\
6&None,\korean{ㄹ},1&&&&&& \\
7&&None,,None&&&&& \\
8&&None,,None&&&None,,None&-1,,2& \\
9&&None,,None&None,,None&None,,None&None,,None&& \\
10&&None,,None&None,,None&None,,None&None,,None&&None,,None \\
11&&None,,None&None,,None&None,,None&None,,None&& \\
12&&None,,None&None,,None&None,,None&None,,None&& \\
13&&None,,None&None,,None&None,,None&None,,None&& \\
14&&None,,None&None,,None&None,,None&&&None,,None \\
15&&&&&&& \\
16&None,,None&&&&&& \\
17&&&&&&& \\
18&None,,None&&&&&& \\
19&&&&&&& \\
20&-1,\korean{ㄹ},None&&&&&& \\
21&&&&&&& \\
22&-1,\korean{ㅇㅝ},2&&&&&& \\
23&&&&&&& \\
24&-1,,None&&&&&& \\
25&-2,\korean{ㄹㄹㅏ},2&&&&&& \\
26&-2,\korean{ㄹㄹ},1&&&&&& \\
27&&&&&&& \\
28&None,,1&&&&&& \\
29&-1,\korean{ㅏ},2&&&&&& \\
30&-1,,1&&&&&& \\
31&&None,,None&&-1,,None&None,,None&& \\
32&&None,,None&&&None,,None&None,,None& \\
33&&None,,None&None,,None&None,,None&None,,None&& \\
34&&None,,None&&&None,,None&-1,\korean{ㅇㅜ},2& \\
35&&None,,None&&&None,,None&-1,\korean{ㅇㅜ},2& \\
36&&None,,None&&&None,,None&-1,,None& \\
37&&None,,None&&&None,,None&-1,,2& \\
38&&None,,None&&&None,,None&-1,,2& \\
39&&None,,None&&&None,,None&-1,,2& \\
40&&None,,None&&&None,,None&-1,,2& \\
41&&None,,None&&-1,,None&None,,None&& \\
42&&None,,None&None,,None&None,,None&None,,None&& \\
43&&None,,None&None,,None&None,,None&None,,None&& \\
44&&None,,None&&&None,,None&None,,None& \\
45&&None,,None&None,,None&None,,None&None,,None&& \\
46&&None,,None&None,,None&None,,None&None,,None&& \\

\Xhline{1.0pt}
\end{tabular}
\end{table}

\begin{table}[ht]
\renewcommand{\arraystretch}{1.2}
\small
\rowcolors{1}{}{Gray}
\begin{tabular}{c|ccc}
\Xhline{1.0pt} 
\rowcolors{1}{}{Gray}
Ending Class  $\triangleright$   & \multirow{2}{*}{22}   & \multirow{2}{*}{23}   & \multirow{2}{*}{24}       
 \\
Verb Class $\triangledown$&&
\\
\Xhline{1.0pt}

1&&& \\
2&&& \\
3&&&None,,1 \\
4&&None,,None& \\
5&&& \\
6&&& \\
7&&& \\
8&&& \\
9&&&None,,1 \\
10&None,,None&& \\
11&&& \\
12&&& \\
13&&& \\
14&None,,None&& \\
15&&& \\
16&&& \\
17&&& \\
18&&& \\
19&&& \\
20&&& \\
21&&& \\
22&&& \\
23&&& \\
24&&& \\
25&&& \\
26&&& \\
27&&& \\
28&&& \\
29&&& \\
30&&& \\
31&&& \\
32&&& \\
33&&& \\
34&&& \\
35&&& \\
36&&& \\
37&&& \\
38&&& \\
39&&& \\
40&&& \\
41&&& \\
42&&& \\
43&&& \\
44&&& \\
45&&& \\
46&&& \\
\Xhline{1.0pt}

\end{tabular}
\end{table}

\end{document}